\documentclass[conference]{IEEEtran}
\IEEEoverridecommandlockouts
\usepackage{cite}
\usepackage{amsmath,amssymb,amsfonts}
\usepackage{algorithmic}
\usepackage{graphicx}
\usepackage{textcomp}
\usepackage{xcolor}
\def\BibTeX{{\rm B\kern-.05em{\sc i\kern-.025em b}\kern-.08em
    T\kern-.1667em\lower.7ex\hbox{E}\kern-.125emX}}
\usepackage{booktabs}
\usepackage{subcaption}
\usepackage{todonotes}
\usepackage{tikz}
\usepackage{url}
\usepackage{comment}
\usepackage{empheq}
\usepackage{multirow}
\usepackage{booktabs} %

\usepackage{xspace}
\newcommand{\h}{\ensuremath{\mathbf{h}}\xspace}
\renewcommand{\L}{\ensuremath{\mathcal{L}}\xspace}
\newcommand{\D}
{\ensuremath{\mathcal{D}}\xspace}
\newcommand{\F}
{\ensuremath{\mathcal{F}}\xspace}

\renewcommand{\S}
{\ensuremath{\mathcal{S}}\xspace}
\newcommand{\Rl}
{\ensuremath{\mathcal{R}^{(l)}}\xspace}
\newcommand{\Sl}
{\ensuremath{\mathcal{S}^{(l)}}\xspace}
\newcommand{\ril}
{\ensuremath{\mathbf{r}_i^{(l)}}\xspace}
\newcommand{\sil}
{\ensuremath{\mathbf{s}_i^{(l)}}\xspace}
\newcommand*\widefbox[1]{\fbox{\hspace{0.5em}#1\hspace{0.5em}}}

\newif\ifproofread

\proofreadtrue

\begin{document}
\title{(Sometimes) Less is More:  Mitigating the Complexity of Rule-based Representation for Interpretable Classification}
\author{
\IEEEauthorblockN{%
Luca Bergamin\IEEEauthorrefmark{1},
Roberto Confalonieri\IEEEauthorrefmark{1}, 
Fabio Aiolli\IEEEauthorrefmark{1}
}
\IEEEauthorblockA{%
\IEEEauthorrefmark{1}\textit{Department of Mathematics 'Tullio Levi-Civita', University of Padova}, Padova, Italy\\
\{bergamin,aiolli\}@math.unipd.it, roberto.confalonieri@unipd.it
}
}

\maketitle

\begin{abstract}
Deep neural networks are widely used in practical applications of AI, however, their inner structure and complexity made them generally not easily interpretable. Model transparency and interpretability are key requirements for multiple scenarios where high performance is not enough to adopt the proposed solution. In this work, a differentiable approximation of $L_0$ regularization is adapted into a logic-based neural network, the Multi-layer Logical Perceptron (MLLP), to study its efficacy in reducing the complexity of its discrete interpretable version, the Concept Rule Set (CRS), while retaining its performance. The results are compared to alternative heuristics like Random Binarization of the network weights, to determine if better results can be achieved when using a less-noisy technique that sparsifies the network based on the loss function instead of a random distribution. {The trade-off between the CRS complexity and its performance is  discussed.}
\end{abstract}

\begin{IEEEkeywords}
Logical Perceptron, Propositional Network, Interpretable Classification
\end{IEEEkeywords}

\section{Introduction}

Advances in
deep learning
have promoted the training of models with a continuously increasing number of parameters in the search for higher performance and newer capabilities, reaching the order of billions in some cases. However, some of these solutions became black-box models incapable of explaining the reasoning behind their decisions~\cite{ALI2023101805}. Some scenarios with critical use cases, such as medicine, law, or finance, demand a higher level of explainability.

Therefore, in the last years, there has been a rise in the number of works, expert groups, and companies that are focusing on transparency and explainability issues \cite{balasubramaniam2023transparency,ALI2023101805}.

Interestingly, there can be a natural synergy in the search for reducing model complexity for efficiency purposes, and the need for transparency in the machine learning field, for which Explainable AI techniques can aid in both reducing model complexity while improving the interpretability of the solutions \cite{sabih2020utilizing
}. However, in some cases, there is a trade-off, for which compression might affect the explainability of the original model, which needs to be assessed and controlled \cite{arazo2024xpression
}. %

There are two important aspects of the interpretability of neural networks. The first one, which is a key notion of transparency as explained in \cite{lipton2018mythos}, is that each part of a model %
should have an intuitive explanation. In the case of neural networks, the use of activation functions in the neurons makes the understanding of the transformation of each input not feasible for humans. A second aspect is the existing trade-off for interpretability of neural networks among faithfulness, understandability, and model performance. Most of the methods in the literature compromise at least one of those requirements, which might not be suitable for highly sensitive scenarios.%

Typically, when explaining a black-box model using a surrogate symbolic model (e.g., ruleset, decision tree), accuracy is often sacrificed for transparency. To address this, Wang et al.~\cite{mllp} proposed a hierarchical rule-based model obtained using a Multi-layer Logical Perceptron network (MLLP), where a rule-based model is learned through backpropagation, and discretized later in a ruleset form.
A key challenge for rule-based models is finding an easily interpretable, concise structure. In this work, we claim that a sparser network naturally leads to simpler rules.
Thus, to achieve higher interpretability, we promote network sparsity through the introduction of a regularization term into the neural network's loss function. Specifically, we apply a differentiable approximation of $L_0$ regularization~\cite{l0} and study its effectiveness in aligning the trained continuous model with its discrete, interpretable version.

The main contributions of this work are:
\begin{itemize}
    \item Reproduction of original experiments from \cite{mllp} in order to verify the results reported and determine a baseline for our approach.

    \item Analysis of the sparsity achieved using $L_0$ regularization adapted to the MLLP framework.

    \item Comparison between the MLLP and the framework adapted to use the $L_0$ regularization, including model performance and complexity analysis.

\end{itemize}

\noindent Our code can be found at \url{github.com/BouncyButton/mllp_l0}. %

\section{Related Work}\label{sec:related_work}

Rule-based models, often presented as decision trees, rule lists, and rule sets, are widely present in the field due to their transparent inner structure, in contrast to other approaches such as Deep Neural Networks (DNN) which are considered black-box models hard to interpret~\cite{ConfalonieriCWB21}. %
Multiple works have explored using backpropagation and/or multilayer structures to learn rule-based models and improve their performance while retaining their transparency \cite{wang2021scalable, beck2021empiricalinvestigationdeepshallow, dierckx_rl-net_2023}. Using differentiable proxies of logic operations is a recently studied topic, due to its easy integration with gradient-based optimization \cite{van_Krieken_2022}.

Wang et al.~\cite{mllp} present a new neural network architecture, the Multi-layer Logical Perceptron (MLLP), which attempts to extract a hierarchical rule set model, using tailored techniques to align the performance between the continuous and discrete solutions. Later, Wang et al.~\cite{wang2021scalable} propose a new classifier, the Rule-based Representation Learner (RRL), that can automatically learn both the continuous and discrete solutions with an improved training method. Its improvement is also due to a different model structure, which considers continuous weights for each rule, boosting therefore the final accuracy of the model, but arguably reducing its overall interpretability.

Model compression for neural networks is a relevant field of study in deep learning since it has been shown that models can be over-parameterized, leading to unnecessary computational resource usage, reduced efficiency, and lower generalization capabilities \cite{sabih2020utilizing}.
Finally, Binary Neural Networks (BNN) are neural networks for which the weights are binary. %
Courbariaux and Bengio~\cite{bnn} propose an efficient method to train BNNs with binary weights and activations,
making use of the straight-through estimator to allow for gradient descent over non-differentiable functions. Even when reducing the computational complexity and improving the transparency, these models are still not considered interpretable due to the high number of non-zero weights.

The aim of regularization within the machine learning field is to reduce the generalization error without increasing its training error~\cite{Goodfellow-et-al-2016}, i.e. to prevent a model from overfitting (which is specially an issue in neural networks due to its generally complex and deep structure), ensuring its performance is aligned for both the training data and new unseen data. Some popular regularization techniques are $L_2$ norm regularization (also known as weight decay) and Dropout regularization. $L_0$ regularization is a type of regularization that imposes a penalty on the objective function directly by the number of non-zero parameters, highly promoting sparsity and improving generalization. Furthermore, it can help speed up inference and training, as those weights that become zero remove some paths in the computational graph. $L_0$ regularization is an NP-hard problem from the computational complexity perspective \cite{l0_nphard}, which is considered difficult to solve. Furthermore, the $L_0$ norm of the weights of a network is not differentiable, so it cannot be incorporated as a regularization term in the loss function.
In \cite{l0}, a tractable, approximate approach to $L_0$ regularization is proposed, with some experimental results that demonstrate its effectiveness in reducing the size of neural networks, such as AlexNet. The authors achieve this by smoothing the expected regularization objective so that network parameters can be zeros while allowing for gradient optimization, making the expected $L_0$ regularization differentiable with respect to the parameters of a given distribution. We argue that the application of this regularization is a great fit for logic-based networks, since sparsity is, in our opinion, a key ingredient to build interpretable classifiers.

\vspace{1em}
\section{Background}\label{sec:background}

\noindent In this section, we introduce the notation and concepts used.%

\subsection{Notation}

In the paper, we make use of the following notation. $J$ is a set of binary features and $C$ is a set of class labels. $\D = \{(\mathbf{x_1}, \mathbf{y_1}),\ldots,(\mathbf{x_n} , \mathbf{y_n})\}$ denotes a training
dataset with $n$ instances, where $\mathbf{x_i} \in \{0,1\}^{|J|}$ is a
binary feature vector and $\mathbf{y_i} \in \{0,1\}^{|C|}$ is a one-hot class label vector. Let $A$ denote the set of binary features, and $a_j \in A$ is the $j$-th binary feature.
We will consider two types of networks: Concept Rule Set and Multi-layer Logical Perceptron. In these networks,
$\Rl$ and $\Sl$ will denote the $l$-th layer of the network when $l$ is odd and even, respectively. These layers are also known as hidden layers. $\S^{(0)}$ denotes the input layer which consists of $|J|$ input neurons. The output layer consists of $|C|$ output neurons. Given $\mathbf{n}_l$ to denote the number of nodes in the $l$-th
layer, $W^{(l)}$ is a $\mathbf{n}_l \times \mathbf{n}_{l-1}$ matrix containing the weights (parameters) of the $l$-th layer and the ($l$-1)-th layer. Each element of $W^{(l)}$ is referred as $w_{i,j}^{(l)}$.
These weights can be learned during training using an optimization algorithm.

A rule $\mathbf{r}$ is a conjunction of one or more Boolean variables $\mathbf{r} = b_1 \wedge \ldots \wedge b_k$.  A rule set $\mathbf{s}$ is a disjunction
of one or more rules $\mathbf{s} = r_1 \vee \ldots \vee r_l$, i.e., $\mathbf{s}$ is a Disjunctive Normal Form (DNF) clause.

\subsection{Concept Rule Set}

\begin{figure}
\centering
\includegraphics[width=1\columnwidth]{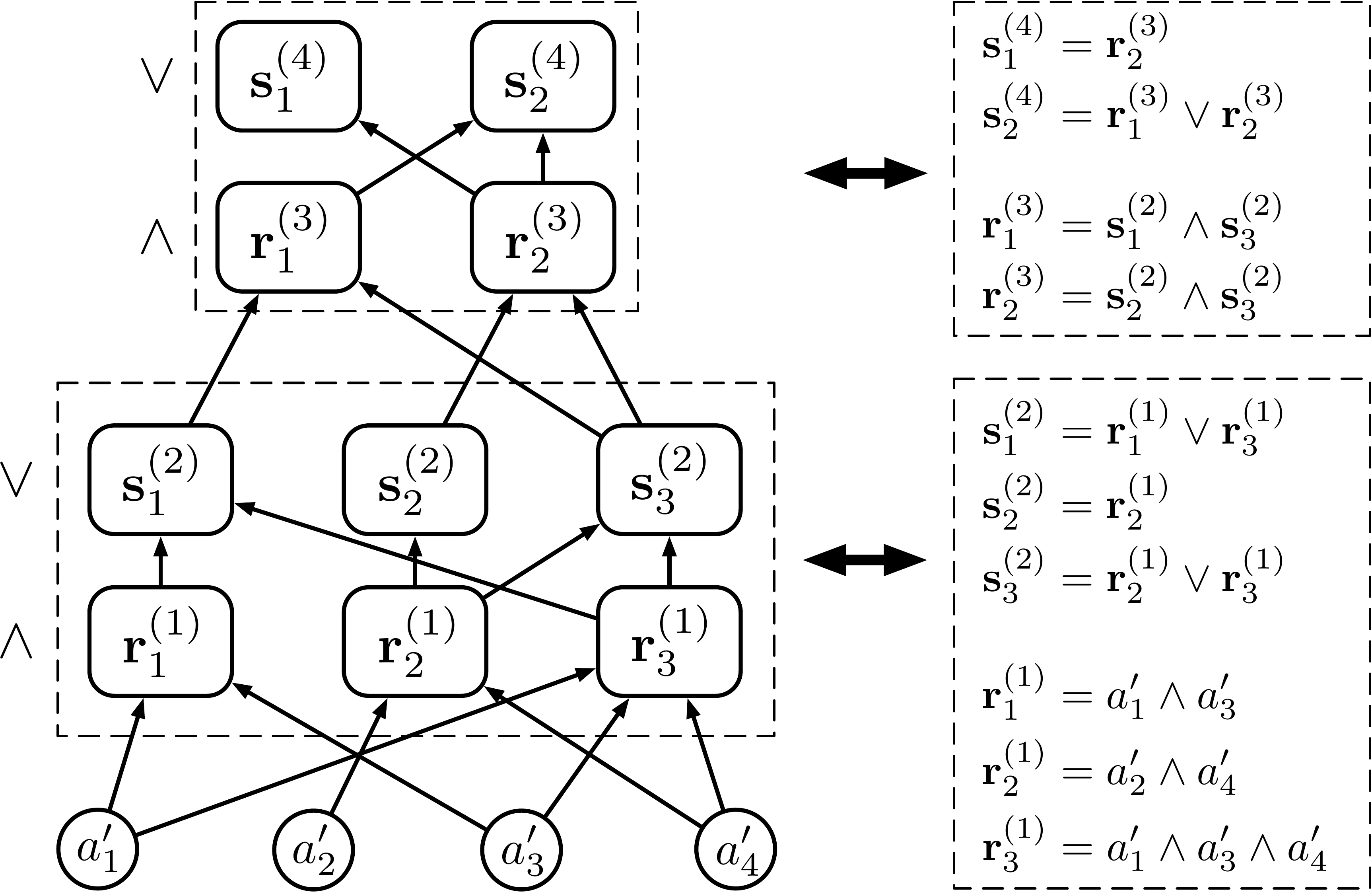}
\caption{Concept Rule Set example taken from~\cite{mllp}. Conjunction layers represent rules and disjunction layers represent rule sets as defined in Eq.~\ref{eq:crs_layers}.}
\label{fig:crs}
\end{figure}

A Concept Rule Set (CRS) is a multi-level hierarchical rule-based model proposed in~\cite{mllp}.
A CRS is an instance of a multi-layer logical perceptron network (see Section~\ref{subsec:mllp}) where each weight in the network is binary, i.e., $w_{i,j}^{(l)} \in \{0,1\}$. $w_{i,j}^{(l)} = 1$ indicates that there exists an edge connecting the $i$-th node in the $l$-th layer to the $j$-th node in the ($l$-1)-th layer, otherwise $w_{i,j}^{(l)} = 0$. Following the notation adopted in~\cite{mllp}, $\mathbf{r}_i^{(l)}$ and $\mathbf{s}_i^{(l)}$ denote the $i$-th node in layer $\Rl$ and $\Sl$, respectively. These nodes are formally defined as follows:
\begin{align}\label{eq:crs_layers}
&\mathbf{r}_i^{(l)}=\bigwedge_{w_{i, j}^{(l)}=1} \mathbf{s}_j^{(l-1)}, \quad \mathbf{s}_i^{(l+1)}=\bigvee_{w_{i, j}^{(l+1)}=1} \mathbf{r}_j^{(l)}
\end{align}
Given the above, node $\ril$  corresponds to a {\em rule},
while node $\sil$ corresponds to a {\em rule set}.
In a CRS with $L$
levels, each $\Rl$ ($l \in\{1,3, \ldots, 2 L-1\}$) is known as a conjunction layer and each $\Sl$ ($l \in\{2,4, \ldots, 2 L\}$) is known as a disjunction layer.
A CRS consists of $2L+1$ layers organized as one input layer followed by pairs of conjunction and disjunction layers. Fig. \ref{fig:crs} represents an example of a CRS  architecture.

By providing each input instance as  values of the input layer $\mathcal{S}^{(0)}$, once trained, a CRS model works as a classifier $\F: \{0, 1\}^{|J|} \rightarrow \{0, 1\}^{|C|}$. The model outputs
the values of nodes in the last disjunction layer $\mathcal{S}^{(2L)}$, where $\mathbf{s}_i^{(2L)} = 1$
indicates that $\F$ classifies the input instance as the $i$-th class label. The representation learned by the $l$-th layer in CRS is a binary vector $\mathbf{h}^{(l)}$:
\begin{align}
\mathbf{h}^{(l)}= \begin{cases}{\left[\mathbf{r}_1^{(l)}, \mathbf{r}_2^{(l)}, \ldots, \mathbf{r}_{\mathbf{n}}^{(l)}\right]^{\top}} & l \in\{1,3, \ldots, 2 L-1\} \\ {\left[\mathbf{s}_1^{(l)}, \mathbf{s}_2^{(l)}, \ldots, \mathbf{s}_{\mathbf{n}}^{(l)}\right]^{\top}} & l \in\{2,4, \ldots, 2 L\}\end{cases}
\end{align}

The value of $\mathbf{h}_i^{(l)}$ is equal to the value of the $i$-th node in the $l$-th layer which corresponds to a rule or a rule set. This discrete and explicit representation makes the model  transparent.

We refer to the {\em complexity} of a CRS model as the total length of all rules. We use $| \mathbf{r}_i^{(l)} |$ and $| \mathbf{s}_i^{(l)} |$ to refer to the number of nodes contained in a rule and rule set, respectively. Then, the complexity of a CRS model ($\mathbf{c}_{\mathcal{F}}$)  is defined as follows:
\begin{equation}\label{eq:complexity}
    \mathbf{c}_{\mathcal{F}} = \sum_{l=1}^{L} \left( \sum_{i=1}^{\mathbf{n}_l} |\mathbf{r}_i^{(2l-1)}| + \sum_{i=1}^{\mathbf{n}_l} |\mathbf{s}_i^{(2l)}| \right)
\end{equation}

Following the example in Fig. \ref{fig:crs}, the complexity $\mathbf{c}_{\mathcal{F}}$ for the represented CRS model would be calculated as $((2 + 2 + 3) + (2 + 1 + 2)) + ((2 + 2) + (1 + 2)) = 19$.

Wang et al.~\cite{mllp} propose the use of an MLLP model (introduced in the next section), a neural network architecture, to search for the discrete solution of the CRS model in the continuous space by using gradient descent over the continuous weights of the MLLP model.  Subsequently, the weights of the network are binarized to transform the MLLP model into a CRS model, resulting in a classifier that ensures both good performance and transparency. For the discrete CRS extraction, a simple method of binarizing the weights using a threshold is applied.

\subsection{Multi-layer Logical Perceptron}\label{subsec:mllp}

The Multi-layer Logical Perceptron (MLLP) is a neural network architecture proposed in~\cite{mllp}, designed in such a way that each of its neurons corresponds to one
node in the CRS.
The main difference with a fully connected Multi-layer Perceptron is the specific design of the activation functions, aiming to replicate the behavior of conjunction and disjunction logical operations.  Another important difference is the presence of a selection mechanism that is used to attend to a subset of its given inputs.

Given the $n$-dimensional vectors \h (a layer input vector) and $\hat{W}_i$ (the weights of a given neuron), and $\hat{w}_{i,j} \in [0,1]$ (the weight of the connection between the input $h_j$ and the neuron $\hat{W}_i$),  the conjunction and disjunction functions  are given by:
\begin{align}
\text{Conj}(\h, \hat{W_i}) &=  \prod_{j=1}^{n} F_c(h_j, \hat{w}_{i,j}), \label{equation:conj} \\
\text{Disj}(\h, \hat{W}_i) &= 1 - \prod_{j=1}^{n} \left(1 - F_d(h_j, \hat{w}_{i,j})\right), \label{equation:disj} \\
F_c(h, w) &= 1 - w(1 - h), \quad
F_d(h, w) = h \cdot w \label{equation:fd}
\end{align}

In $\text{Conj}(\h, \hat{W_i})$, the conjunction operation is obtained by multiplying many values between 0 and 1 together. For $\text{Disj}(\h, \hat{W}_i)$, the disjunction is computed with a similar operation, applying Morgan's law by negating both the inputs and the outputs of the function. To implement this negation, $N(x) = 1-x$ is used.
Then, $F_c$ and $F_d$ act as a selection mechanism. By turning a weight $w$ to zero, $F_c$ can learn to output 1, regardless of its input, thus leaving the subsequent conjunction operation unaffected. A similar mechanism is implemented for $F_d$, making sure to output zero instead. Notice that when \h and $\hat{W}_i$ are both binary vectors, Eq.~\ref{equation:conj} and \ref{equation:disj} reduce to the conjunction and disjunction of a subset of the elements in $\h$, respectively. The subset considers only the elements $\h_j$ where $\hat{w}_{i,j}$ is 1.

The conjunction and disjunction functions are applied to the
neurons in the $l$-th layer of an MLLP as follows:
\begin{align}
\hat{\mathbf{r}}_i^{(l)}=\text{Conj}\left(\hat{\mathbf{s}}^{(l-1)}, \hat{W}_i^{(l)}\right), l \in\{1,3, \ldots, 2 L-1\}\label{equation:mllp_conj} \\
\hat{\mathbf{s}}_i^{(l)}=\text{Disj}\left(\hat{\mathbf{r}}^{(l-1)}, \hat{W}_i^{(l)}\right), l \in\{2,4, \ldots, 2 L\}\label{equation:mlpp_disj}
\end{align}
where $\hat{\mathbf{r}}_i^{(l)}$ and $\hat{\mathbf{s}}_i^{(l)}$ are  neurons in the $l$-th layer of the MLLP, and $\hat{W}^{(l)}$ is a $\mathbf{n}_{l} \times \mathbf{n}_{l-1}$ weight matrix.

The weights of the MLLP in the network need to be constrained in the range $[0,1]$ to ensure the proper functioning of the conjunction and disjunction activation functions. To this end, the weights of a given layer $l$ are constrained using a clip function:%
as follows:
\begin{equation}
\text{Clip}(\hat{w}_{i,j}^{(l)}) = \max\left(0, \min \left(1, \hat{w}_{i,j}^{(l)}\right)\right)
\label{equation:clip}
\end{equation}

\noindent Given the encoding above, the MLLP network can function identically to the corresponding CRS when the weights are set to the same discrete values, while still remaining differentiable.

For training the MLLP, a loss function is defined using the Minimum Squared Error (MSE) criterion. This criterion is considered to be appropriate by the authors of MLLP to account for the difference between the continuous output and the one-hot encoded label vector, which the authors believe can be beneficial during the CRS extraction.

Let $\mathcal{\hat{F}}$ be the MLLP model and $\hat{W}$ be the weights to be learned by the network. The MLLP loss function is given by:
\begin{equation}
\L(\hat{W}) = \frac{1}{N} \sum_{i=1}^{N} \textit{MSE}\left( \hat{\mathcal{F}}(\text{x}_i, \hat{W}),\text{y}_i \right) + \lambda \sum_{l=1}^{2L} \left\| \hat{W}^{(l)} \right\|_2^2
\label{equation:loss_mllp}
\end{equation}

\noindent The second term in the rhs of Eq.~\ref{equation:loss_mllp} is the $L_2$ regularization
term. This term
is included to shrink the MLLP weights and it induces a simpler CRS model.

After training the continuous model (MLLP), a discrete and interpretable model (CRS) can be extracted. Its behavior is not guaranteed to follow the continuous version, and the authors observe a drop in performance. This problem arises with the application of conjunction and disjunction operations over continuous weights, as the MLLP weights can be in the range [0,1]. To mitigate this issue, the authors propose a training method based on Random Binarization (RB).

The RB method selects some of the weights during training using a random binarization rate ($\mathcal{P} \in [0,1]$), which is the probability of a weight to be binarized. The binarized weights are frozen and the forward (predict) and backpropagation  are performed. After a given number of steps, the binarized weights are restored and a new set of random weights is binarized (in the experiments, the RB operation is applied every epoch). Therefore, during training, the MLLP model has a more aligned behaviour to that of the CRS, what ensures a closer performance between the MLLP and CRS models after the binarization conversion of the continuous solution. The RB operation over the weights of each layer, where $\mathbb{I}$ represents the function that determines the binary output of an input based on a threshold $\mathcal{T}$ (in the experiments, $\mathcal{T} = 0.5$) and $M^{(l)}$ represents a $\mathbf{n}_l \times \mathbf{n}_{l-1}$ random binarization mask applied to the $l$-th layer of the MLLP, can be formalized as follows:
\begin{align}
    Bin(w, \mathcal{T})&=\mathbb{I}(w > \mathcal{T}), \label{equation:binarize} \\
    m_{i,j}^{(l)} \in \{0,1\}, \quad  m_{i,j}^{(l)}&=\mathbb{I}(p<\mathcal{P}), \quad p \sim \textit{U}(0, 1), \nonumber \\
    \tilde{w}_{i,j}^{(l)} = RB(\hat{w}_{i,j}^{(l)}, m_{i,j}^{(l)}, \mathcal{T})
    &=
    \begin{cases}
    \text{$\hat{w}_{i,j}^{(l)}$}& \text{$m_{i,j}^{(l)}=0$},\nonumber \\
    \text{$Bin(\hat{w}_{i,j}^{(l)}, \mathcal{T})$}& \text{$m_{i,j}^{(l)}=1$}
    \end{cases}
\end{align}

\noindent $\tilde{W}_{i}^{(l)}$ denotes the weights of the $i$-th neuron in the $l$-th layer of the MLLP after RB. These weights  replace $\hat{W}_{i}^{(l)}$ in Eq.~\ref{equation:mllp_conj}~and~\ref{equation:mlpp_disj}. The behavior of the RB method is similar to that of the Dropout regularization \cite{dropout}, %
addressing overfitting. %

Once the MLLP model has been trained and converted into a CRS model, additional steps can be performed to reduce the size of the model and benefit its interpretability, such as detecting dead nodes and redundant rules. %

\subsection{$L_0$ regularization Approximation}

In this work, we focus on the $L_0$ regularization approximation proposed in \cite{l0}. The authors propose a set of gates that determine whether a parameter or group of parameters of the network (e.g., all the weights associated to a given input neuron of a layer of the network) are active (i.e., value greater than 0) or inactive (i.e., value equal to 0). The values $z$ of those gates can be drawn from a distribution such as the binary Bernoulli. However, this distribution needs to be smoothed in order to be differentiable. Given $s$ as a continuous random variable with a distribution $q(s)$ and parameters $\phi$, rectified to be in the interval [0,1], the probability of the gate being active can be calculated using its cumulative distribution function (CDF) so that:
\begin{align}
    s \sim q(s | \phi), \quad &z = \min(1, \max(0, s)), \\
    q(z \neq 0 | \phi) &= 1 - Q(s \leq 0| \phi) \nonumber
\end{align}

The selected distribution $q$ proposed in this work is the \textit{hard concrete} distribution, obtained by stretching a \textit{binary concrete} distribution~\cite{conc_dist,conc_dist_2
}, closely tied to the Bernoulli distribution, and rectifying each sampled value $z$ using a hard-sigmoid to constrain its values in the range $[0,1]$. The parameters $\phi$ for this distribution are log $\alpha$ as the {\em location} and $\beta$ as the {\em temperature} (to control the degree of recovery of the original Bernoulli distribution). $\zeta$ and $\gamma$ determine the stretching of the original distribution. The \textit{hard concrete} distribution can be formalized as follows:
\begin{align}
u &\sim \textit{Uniform}(0,1), \label{equation:uniform_rep_trick}\\
s &= \text{Sigmoid}\left(\frac{\log u - \log(1 - u) + \log \alpha}{\beta}\right), \label{equation:cdf_rep_trick} \nonumber \\
\bar{s} &= s(\zeta - \gamma) + \gamma, \quad z = \min\left(1, \max(0, \bar{s})\right) \nonumber
\end{align}
The nature of the \textit{binary concrete} distribution allows for the reparametrization trick~\cite{rep_trick_vae%
}, which prevents the introduction of randomness in the gradient descent process when involving sampling from the learned distribution (as expressed in Eq. \ref{equation:uniform_rep_trick} and \ref{equation:cdf_rep_trick}). The authors also propose \textit{group sparsity} instead of \textit{parameter sparsity} as a means to achieve network speedups. In their approach, the \textit{group sparsity} is presented as input neuron sparsity in the case of fully connected layers. %

\begin{align}
\hat{\mathcal{R}}(\hat{W}, \phi) &=\mathcal{L}_E(\hat{W}) + \lambda \mathcal{L}_{L0}(\phi) + \lambda' \mathcal{L}_{L2}(\phi), \\
\mathcal{L}_E(\hat{W}) &= \frac{1}{K}\sum_{k=1}^{K}\bigg(\frac{1}{N}\bigg(\sum_{i=1}^{N}\mathcal{L}\big(h(\text{x}_i, \hat{W}\times \text{z}^{(k)}), \text{y}_i\big)\bigg)\bigg), \nonumber \\
\mathcal{L}_{L0}(\phi) &=
\sum_{g=1}^{|G|}|g|\bigg(1 - Q(s_g \leq 0|\phi_g)\bigg), \nonumber \\
\mathcal{L}_{L2}(\phi) &= \sum_{g=1}^{|G|}\bigg(\big(1 - Q(s_g \leq 0|\phi_g)\big)\sum_{j=1}^{|g|}\hat{W}^2_j\bigg) \nonumber
\end{align}

where $|G|$ corresponds to the number of groups and $|g|$ corresponds to the number of parameters of group $g$. $\mathcal{L}_E$ represents the loss criterion.
The probability of a gate $g$ being active under the \textit{hard concrete} distribution for the proposed \textit{grouped sparsity} can be expressed as:
\begin{equation}
\bigg(1 - Q(s_g \leq 0|\phi_g)\bigg) = \text{Sigmoid}\big(\log\alpha_g - \beta\log\frac{-\gamma}{\zeta}\big)
\end{equation}

During test, the deterministic values of the gates are obtained with the following estimator:
\begin{align}
	\hat{z} = \min(1, \max(0, \text{Sigmoid}(\log\alpha)(\zeta - \gamma) + \gamma)) \label{equation:test_gate}
\end{align}

\section{Integrating $L_0$ Regularization into MLLP}\label{sec:proposal}

The original implementation of the MLLP architecture employs the RB method to address the misalignment between the continuous and discrete solutions, while addressing overfitting similarly to Dropout regularization. Although the CRS model learned through the MLLP  achieves high transparency and performance, the resulting solutions can still be complex due to the large number of rules involved. {Our proposal:}

\begin{itemize}
    \item {\bf {Induces sparsity directly}:} $L_0$ regularization directly promotes sparse solutions based on its presence in the loss function. Instead, dropout regularization mainly improves robustness by increasing redundancy, moving away from sparser models. Instead, using Random Binarization does not explicitly take into account the sparsity of the network in the loss function. 
    \item {\bf {Performs feature selection automatically}:} those input neurons that are not relevant can be ignored during the training process.
    \item {\bf {Is more interpretable}:} with the $L_0$ regularization technique, sparser models are preferred.%
    \item {\bf {Can be more efficient}:} training sparse models can be more efficient, by excluding null weights from the computational graph.
\end{itemize}

\subsection{Adaptation of $L_0$ regularization to MLLP framework}

\noindent %
To start, the MLLP network needs to include the locations $\log$ $\alpha$ of a \textit{hard concrete distribution} for each gate as trainable parameters. In our implementation, we follow the same approach as \cite{l0}\footnote{\url{https://github.com/AMLab-Amsterdam/L0_regularization}}, where each gate represents the activation or deactivation of an input neuron, coined as \textit{group sparsity}. These values are initialized by sampling a normal distribution determined by a dropout rate. Furthermore, we use a separate dropout rate for only the input layer only. %

\begin{figure}[!t]
\centering
\includegraphics[width=.75\columnwidth]{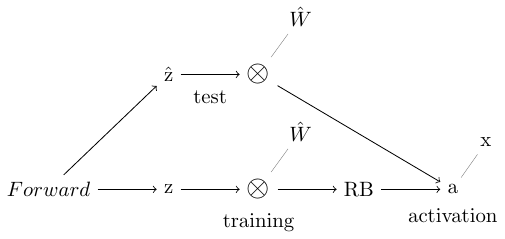}
\caption{Schema of the forward operation of each conjunction and disjunction layer during training and test for the $L_0$ regularization adaptation to the MLLP framework.}
    \vspace{-1.25em}

\label{fig:schema_forward_l0}
\end{figure}

When performing the continuous \textit{forward} operation, the behavior differs during training and testing. During training, the value $z$ of each gate is sampled from the \textit{hard concrete} distribution, using the reparametrization trick by sampling a uniform distribution. Then, the weights are multiplied with the value of its corresponding $L_0$ gate and randomly binarized (when $\mathcal{P} > 0$) before applying the conjunction or disjunction activation functions. During testing, the value of $\hat{z}$ is obtained with a deterministic operation using the expression in Eq. \ref{equation:test_gate}. Fig. \ref{fig:schema_forward_l0} summarizes the operations performed during the forward pass. In the case of the \textit{binarized forward} operation, i.e., the forward operation performed when the MLLP model is converted into a CRS model, a new threshold $\mathcal{T}'$ is included to binarize the $\hat{z}$ value of each $L_0$ gate, following Eq. \ref{equation:binarize}.

In the same way as in the implementation in \cite{l0}, we used only one sample of a gate $z$ for each minibatch during training, assuming this can result in a larger variance in the gradients. We can then define the updated loss function of MLLP which includes $L_0$ regularization as follows:
\begin{align}
\L'(\hat{W},\phi)&= \L_E'(\hat{W}) + \lambda \mathcal{L}_{L0}(\phi) + \lambda' \mathcal{L}_{L2}(\phi), \\
\L_E'(\hat{W}) &= \frac{1}{N}\bigg(\sum_{i=1}^{N}\textit{MSE}\left(\hat{\mathcal{F}}(\text{x}_i, \hat{W}\times \text{z}), \text{y}_i\right)\bigg)
\label{equation:loss_mllp_l0}
\end{align}

Finally, since the original implementation of the $L_0$ regularization is applied to classical neural networks, some minor adaptations were needed. First,
since the weights in the MLLP network are constrained between 0 and 1, the Kaiming initialization strategy outlined in \cite{l0} would not make sense. Therefore, the initialization is left in the same way as in \cite{mllp}, using \textit{Uniform}(0, 0.1). Second,
MLLP networks do not have any bias terms. Therefore, the bias was considered neither in the neurons nor as locations $\log$ $\alpha$ of the $L_0$ gates.

\section{Evaluation}\label{sec:evaluation}

\noindent In this section, we evaluate the proposed solution as follows:

\begin{itemize}
    \item Reproduction of original experiments from \cite{mllp}. We verify the results reported and establish a baseline.

\item Comparison between the baseline and our approach in terms of CRS and MLLP predictions, as well as the complexity %
of the CRS model.
    \item Analysis of the sparsity achieved with $L_0$ regularization.
    \item {Analysis of the trade-off between the complexity of the CRS model and its performance.}

\end{itemize}

\subsection{Experimental Settings}
The hyperparameter settings are extracted from the Experimental section of~\cite{mllp}, namely batch size of 128, 400 epochs, learning rate of $5\times10^{-3}$, learning rate decay of 0.75 every 100 epochs, weight decay ($\lambda'$) of $10^{-8}$, $\mathcal{T}$ of 0.5, and update of randomly binarized weights every epoch (when applied). 5-cross validation is adopted for more reliable results. For analyzing the performance, the F1 score (Macro) is the chosen metric due to dataset imbalance. The $L_0$ settings are $\lambda=0.001$, $\mathcal{T}'=0.5$, and $\beta=2/3$, as reported by \cite{conc_dist}. The $\lambda$ hyperparameter is to be understood as a value to tune the degree of sparsity of the network.

We consider the \textit{connect-4} dataset from the UCI machine learning repository\footnote{\url{https://archive.ics.uci.edu}} as our benchmark, as it is the only one with an in-depth presentation in \cite{mllp}. Due to the stochasticity of the training process, we expect to find slightly different results. This dataset is relevant for rule-learning models \cite{BERGAMIN2025128699} due to its considerable size (having 67557 instances), and its difficulty: despite being a deterministic game, i.e., data contains no noise, classifiers struggle to achieve high F1-scores.

The architecture of each model is reported using the number of nodes for each hidden layer (e.g., 64 $\times$ 3 represents a model with three hidden layers with 64 nodes each).

\begin{figure}[!t]
    \centering
    \includegraphics[width=\columnwidth, trim=0 0 0 12, clip]{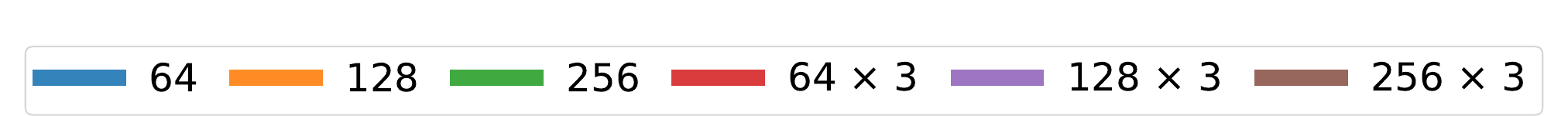}
\begin{subfigure}[b]{0.49\columnwidth}
        \centering
        \includegraphics[width=\linewidth, trim=0 0 0 0, clip]{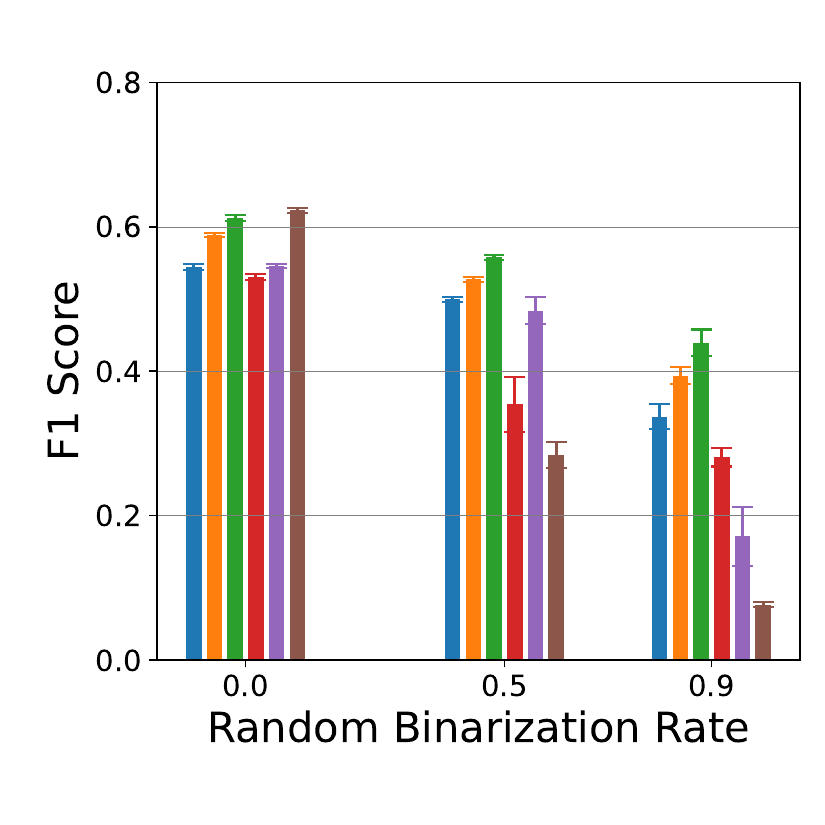}
\vspace{-2em}  
\caption{Baseline}
        \label{fig:repl_connect_4_f1_mllp}
    \end{subfigure}
    \hfill
    \begin{subfigure}[b]{0.49\columnwidth}
        \centering
        \includegraphics[width=\linewidth, trim=0 0 0 0, clip]{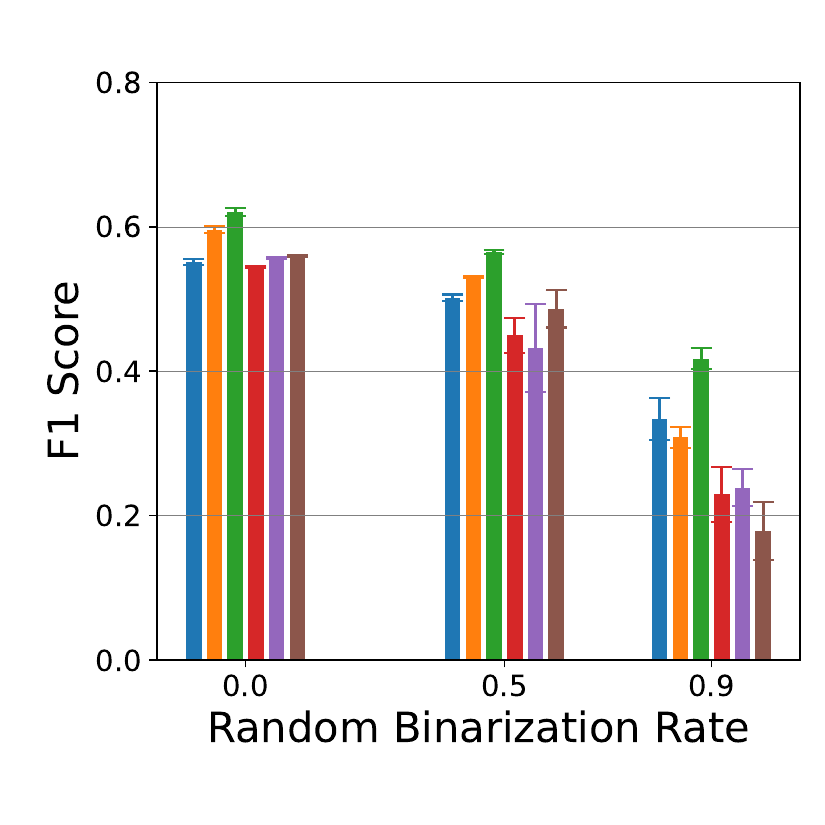}
        \vspace{-2em}  
        \caption{$L_0$ regularization}
        \label{fig:l0_connect_4_f1_mllp}
    \end{subfigure}
    \caption{(a) Replicated \textit{connect-4} results for MLLP models. (b) \textit{connect-4} results including $L_0$ regularization.}
    \label{fig:repl_and_l0_connect_4_f1_mllp}
\end{figure}

\begin{figure}[!]
    \centering
    \vspace{-1em} 
    \includegraphics[width=\columnwidth, trim=0 0 0 12, clip]{IJCNN2025/active-weights-legend.pdf}

\vspace{1em}

\begin{subfigure}[b]{0.49\columnwidth}
        \centering
        \includegraphics[width=\linewidth, trim=0 38 0 34, clip]{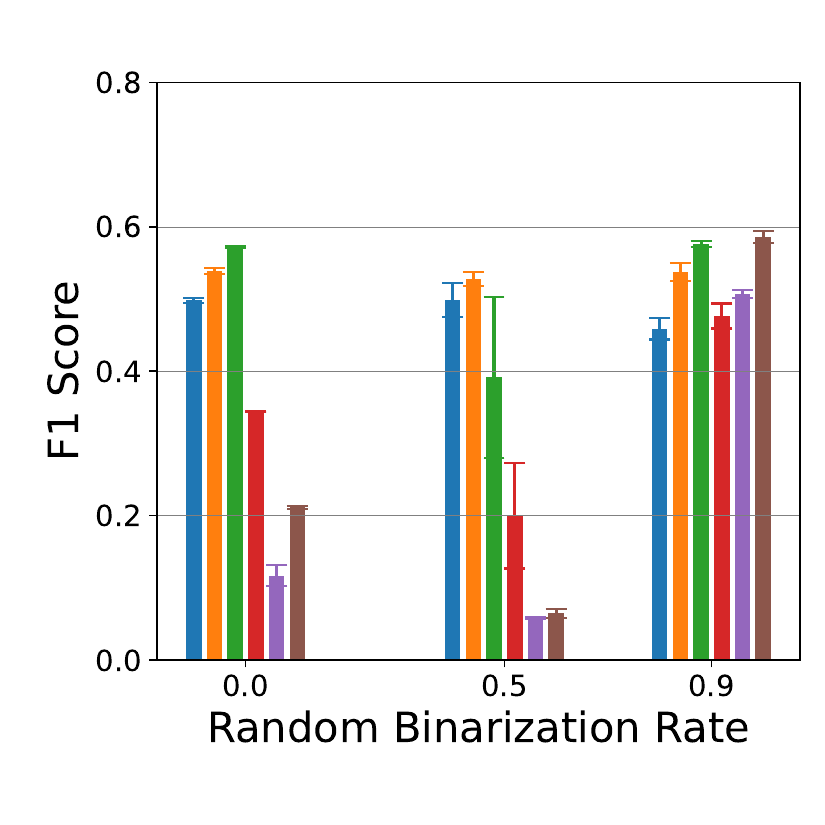}
        \caption{Baseline}
        \label{fig:repl_connect_4_f1_crs}
    \end{subfigure}
    \hfill
    \begin{subfigure}[b]{0.49\columnwidth}
        \centering
        \includegraphics[width=\linewidth, trim=0 38 0 34, clip]{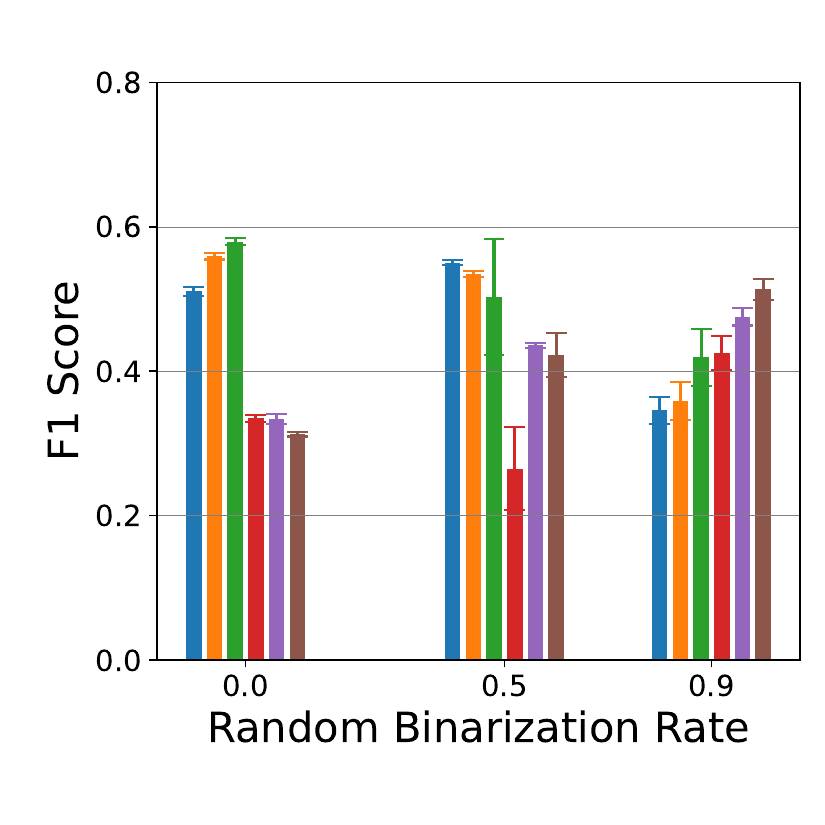}
        \caption{$L_0$ regularization}
        \label{fig:l0_connect_4_f1_crs}
    \end{subfigure}
    \caption{(a) Replicated \textit{connect-4} results for CRS models. (b) \textit{connect-4} results including $L_0$ regularization.}
    \label{fig:repl_and_l0_connect_4_f1_crs}

    \vspace{-1.25em}
\end{figure}

\noindent We replicate the original experiments %
using $\mathcal{P} \in \{0, 0.5, 0.9\}$. 

\subsection{Reproduction of original experiments}

Figs. \ref{fig:repl_and_l0_connect_4_f1_mllp}a and \ref{fig:repl_and_l0_connect_4_f1_crs}a aggregate the reproduction results for multiple values of $\mathcal{P}$ and multiple model structures that range in different depths and widths, displaying the average F1 score and standard deviation %
for the MLLP and CRS models. Focusing on Fig. \ref{fig:repl_and_l0_connect_4_f1_mllp}a, the RB method seems to fail to preserve the performance of the MLLP models, especially with greater values of $\mathcal{P}$. Results for MLLP performance using RB are not reported in \cite{mllp}, despite the use of the RB method that is said to align the performances of both MLLP and CRS models. %
Moving to Fig. \ref{fig:repl_and_l0_connect_4_f1_crs}a, we observe the similarity for both replicated and original results (see Fig. 3 in \cite{mllp}) between the trends for shallow models compared to those of deeper models. In both the original and replicated results, the value of $\mathcal{P}$ does not significantly influence the results of shallow models, whereas it is more influential for deeper models. {This is in line with the intuition that deeper networks can generalize better when the learning is  distributed over the network.}

\subsection{Performance comparison}
Fig. \ref{fig:repl_and_l0_connect_4_f1_mllp} shows the average F1 scores for both the replicated results of the MLLP from \cite{mllp} and our proposal. %
Results are similar or considerably improved with the inclusion of $L_0$ regularization for %
$\mathcal{P} \le 0.5$. However, when increasing $\mathcal{P}$ to 0.9, scores without $L_0$ regularization are generally higher, with some exceptions for deeper models. In both cases, the greater the value of $\mathcal{P}$, the worse the performance of the MLLP models, with a similar trend of deterioration.

Conversely, Fig. \ref{fig:repl_and_l0_connect_4_f1_crs} shows the average F1 scores for both the replicated results of the CRS from \cite{mllp} and our proposal. For $\mathcal{P} \le 0.5$ we mainly observe that performance is mostly aligned or improved when introducing $L_0$ regularization, especially in deeper models. When focusing on $\mathcal{P}=0.9$, CRS models without $L_0$ regularization perform slightly better, especially for shallow models, which %
{experience a decrease in performance when $L_0$ regularization is applied.} 

These results show some limited evidence that $L_0$ regularization can be a valid substitute for RB, and a combination of the two techniques can boost performance. Combining $L_0$ and RB with an excessive $\mathcal{P}$ does not help to improve results, possibly due to an excessive sparsification of the network.

\begin{figure}[!t]
\includegraphics[width=\columnwidth, trim=0 0 0 12, clip]{IJCNN2025/active-weights-legend.pdf}
    \centering
    \vspace{-1em}
    \begin{subfigure}[b]{0.49\columnwidth}
        \centering
        \includegraphics[width=\linewidth, trim=0 0 0 0, clip]{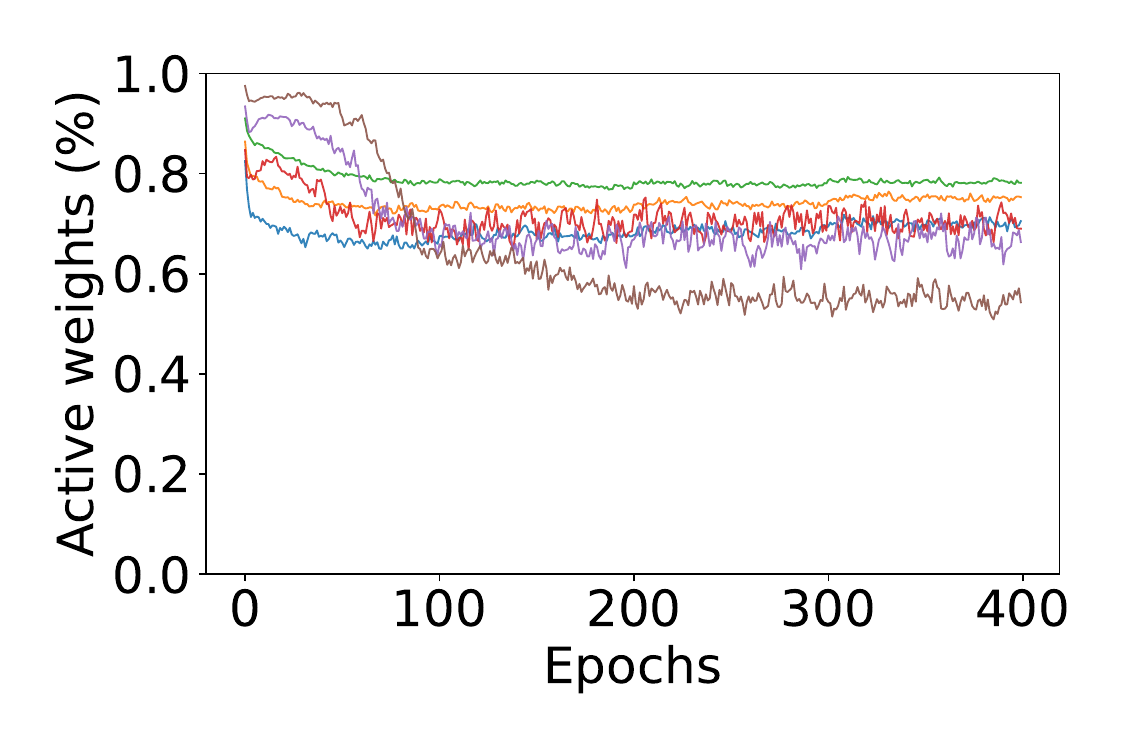}
        \vspace{-2em}
\caption{Baseline}
        \label{fig:repl_connect_4_total_active_weights_mllp}
    \end{subfigure}
    \hfill
    \begin{subfigure}[b]{0.49\columnwidth}
        \centering
        \includegraphics[width=\linewidth, trim=0 0 0 0, clip]{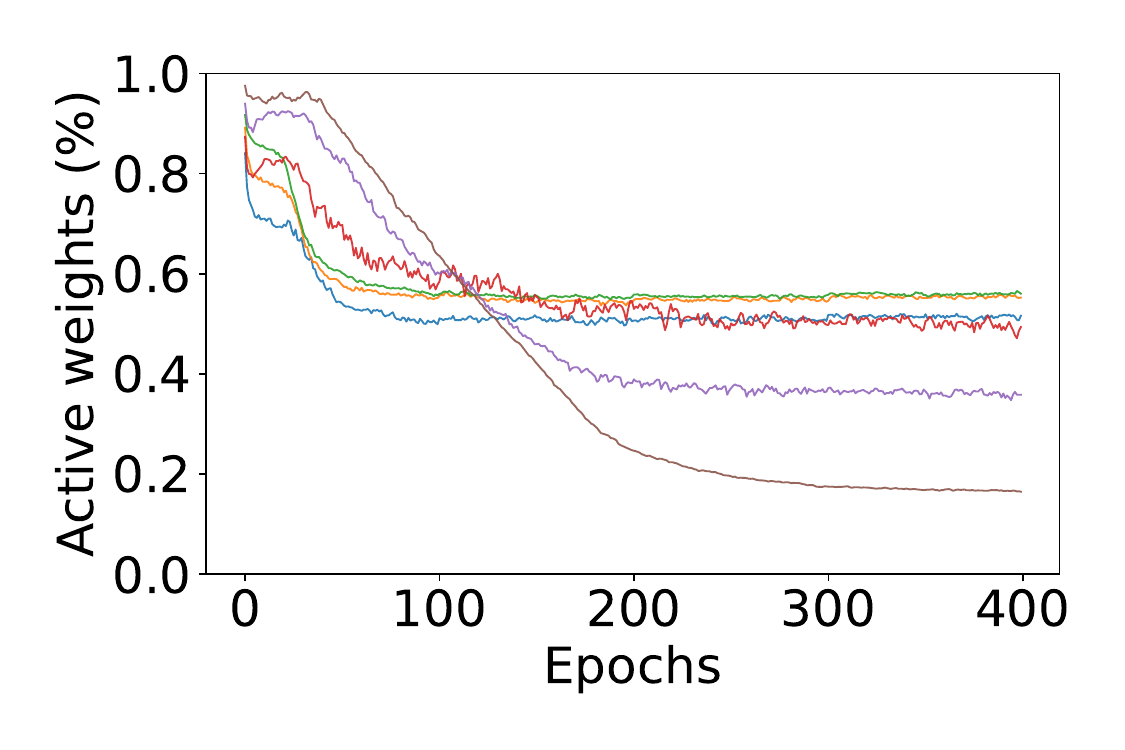}
        \vspace{-2em}

        \caption{$L_0$ regularization}
        \label{fig:l0_connect_4_total_active_weights_perc_mllp}
    \end{subfigure}
    \vspace{0.1em}

    \caption{(a) Active weights during training for \textit{connect-4}. (b) Active weights during training with $L_0$ for \textit{connect-4}.
    \vspace{-1.25em}
    }
    \label{fig:repl_and_l0_total_active_weights_perc_mllp}
\end{figure}

\subsection{Sparsity analysis} Fig. \ref{fig:repl_and_l0_total_active_weights_perc_mllp} shows the evolution of active weights 
of the MLLP models during the training with the \textit{connect-4} dataset, for both the replicated results and our proposal. An active weight is a weight $w$ which has a value greater than 0. 
In the case of the models including $L_0$ regularization, the active weights are those for which the product of the value of the weight $w$ with its corresponding $L_0$ gate $z$ is greater than 0. We observe that the regularized models lead to %
sparser solutions compared to the baseline, which quickly reaches a plateau. Furthermore, the models with a greater number of weights are highly sparse compared to simpler models in terms of active weights. This is more evident in Fig. \ref{fig:repl_and_l0_total_active_weights_perc_mllp}b, where it is shown how drastically the active weights are reduced in the case of the bigger models.

\begin{figure}[!t]
    \centering
        \vspace{-0.5em}
        \includegraphics[width=\columnwidth, trim=0 0 0 12, clip]{IJCNN2025/active-weights-legend.pdf}

\vspace{1em}

    \begin{subfigure}[b]{0.49\columnwidth}
        \centering
        \includegraphics[width=\linewidth, trim=0 38 0 34, clip]{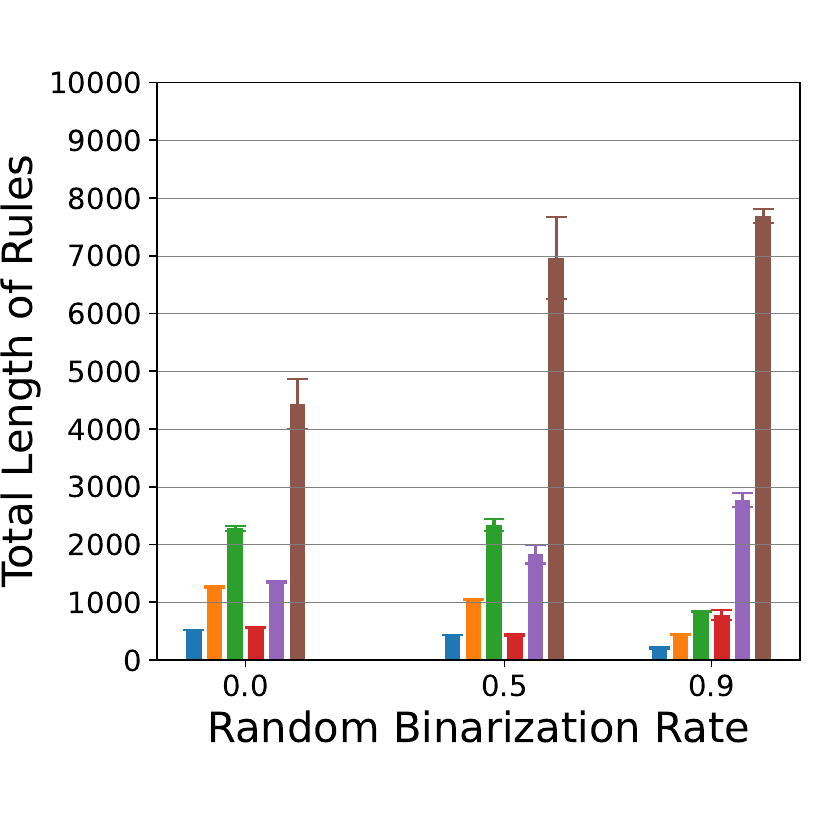}
            \vspace{-1em}

\caption{Baseline}
        \label{fig:repl_connect_4_total_len_rules_crs}
    \end{subfigure}
    \hfill
    \begin{subfigure}[b]{0.49\columnwidth}
        \centering
        \includegraphics[width=\linewidth, trim=0 38 0 34, clip]{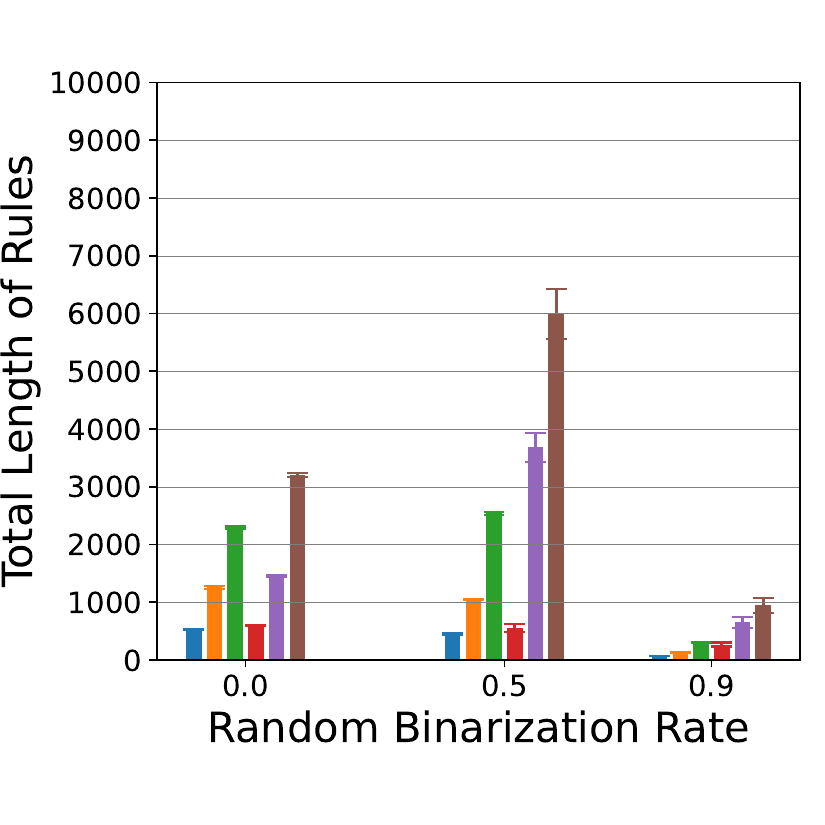}
            \vspace{-1em}

        \caption{$L_0$ regularization}
        \label{fig:l0_connect_4_total_len_rules_crs}
    \end{subfigure}
    \caption{(a) Number of literals in CRS models for \textit{connect-4}. (b) Number of literals in CRS models with $L_0$ for \textit{connect-4}.
    \vspace{-1.25em}
    }
    \label{fig:repl_and_l0_connect_4_total_len_rules_crs}
\end{figure}

\begin{table}[!b]
    \vspace{-1.25em}
\centering
  \caption{Quantitative description of the datasets selected.}
  \label{tab:datasets}
\resizebox{0.8\columnwidth}{!}{%
  \begin{tabular}{lrrrr}
    \toprule
    \textbf{Name} & \textbf{Instances} & \textbf{Classes}	& \textbf{Orig. features}	& \textbf{Bin. features} \\
    \midrule
    \midrule 
adult & 32561 & 2 & 14 & 155\\ 
bank & 45211 & 2 & 16 & 88\\ 
connect-4 & 67557 & 3 & 42 & 126\\ 
letRecog & 20000 & 26 & 16 & 155\\ 
magic04 & 19020 & 2 & 10 & 79\\ 
mush. & 8124 & 2 & 22 & 117\\ 
wine & 178 & 3 & 13 & 37\\
  \bottomrule
\end{tabular}
}
\end{table}

\pagebreak
\subsection{Model complexity}

Fig. \ref{fig:repl_and_l0_connect_4_total_len_rules_crs} illustrates the complexity measure of a CRS model (Eq.~\ref{eq:complexity}), defined as the total rule length (or the number of literals), for both the replicated results from \cite{mllp} and our proposed models. When RB is not applied ($\mathcal{P} = 0$), the total rule length is generally similar in both cases, with the exception of the largest model, whose complexity is considerably reduced. For $\mathcal{P} = 0.5$, the complexity of the $128 \times 3$ model increases under  $L_0$ regularization, which might be related to the drastic increase in performance it promotes. In contrast, the largest model not only shows a substantial performance gain but also achieves a considerable reduction in complexity. Therefore, simplifying a model can, in some cases,  lead to better performance. 
For $\mathcal{P} = 0.9$, $L_0$ regularization helps reduce the complexity of almost every model, especially of the largest one, where the total rule length is reduced by approximately eight times. This reduction might be related to the slight decrease in performance observed.

Finally, it is worth noting that the overall number of rules and literals is large in all models. This seems to suggest that the problem is not easily solvable with propositional rules, thereby necessitating more complex classifiers. This does not mean that propositional rules are insufficient, but their number could substantially grow, since DNF formulas are powerful enough to describe any classification task on a binary dataset. 

\vspace{-1.75em}

\begin{figure}[t!]
    \centering
    \vspace{-0em}
    \begin{subfigure}[b]{0.45\columnwidth}
        \centering
        \includegraphics[width=\linewidth, trim=20 0 357 34, clip]{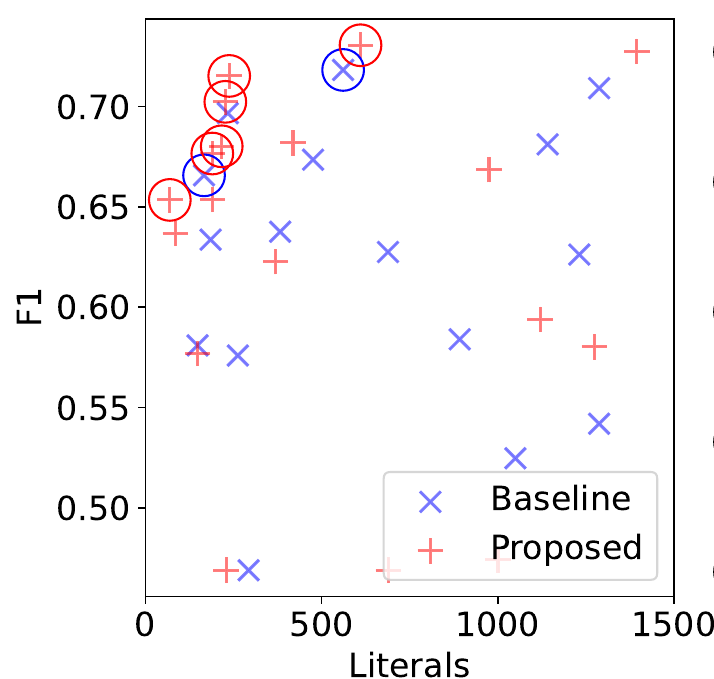}
        \caption{bank-marketing}
        \label{fig:bank}
    \end{subfigure}
    \hfill
    \begin{subfigure}[b]{0.45\columnwidth}
        \centering
        \includegraphics[width=\linewidth, trim=363 0 20 34, clip]{pareto.pdf}
        \caption{letRecog}
        \label{fig:letrecog}
    \end{subfigure}
    \caption{{Comparison of model complexity vs. performance of the CRS classifiers for both models. Best models are circled, highlighting the Pareto frontier, i.e., models which cannot improve their F1 score without having more literals.}
    }
    \vspace{-0.75em}

    \label{fig:rulesvsf1}
\end{figure}

\vspace{1.5em}

\subsection{Trade-off between model complexity and performance}

To further investigate how our proposal influences the resulting models, we performed additional experiments on other UCI datasets, as described in Tab.~\ref{tab:datasets}. Our results, presented in Tab.~\ref{tab:table}, sometimes reveal a slight decrease in performance that is associated with a substantial reduction in model complexity. This outcome is consistent with the regularized loss objective, which may favor simpler models with somewhat lower performance over a more complex,  unregularized alternative. To examine this trade-off between model complexity and performance more closely, we compare both techniques using two different datasets (Fig.~\ref{fig:rulesvsf1}).  In Fig.~\ref{fig:bank} regularized models can surpass the performance of unregularized models while remaining smaller and more efficient. Conversely, Fig.~\ref{fig:letrecog} illustrates that models can become too small--using too few rules--causing a noticeable drop in performance. {In these cases, decreasing the $\lambda$ hyperparameter can help mitigate the issue. A more thorough exploration of tuning $\lambda$ is left for future work.}

{Finally, we present in Tab.~\ref{tab:example} a small example where we show the extracted CRS models for a specific dataset. We observe that the baseline model creates rules that are too specific, due to a larger number of literals. Instead, our model removes potentially redundant features in a data-driven manner, arguably improving generalization.}

\begin{table}[b!]
\centering
    \vspace{-0.75em}

\caption{Mean F1 score of the CRS classifiers and mean number of literals of the best model selected via 5-fold cross-validation. We also report their best $\mathcal{P}$ and architecture.}
\resizebox{0.95\columnwidth}{!}{%
\begin{tabular}{lrrrrrrrr}
\toprule
& \multicolumn{4}{c}{\textbf{Baseline}} & \multicolumn{4}{c}{\textbf{Proposed}} \\
\cmidrule(lr){2-5} \cmidrule(lr){6-9}
\textbf{Name} & \textbf{F1} & \textbf{Compl.} & $\mathcal{P}$ & \textbf{Arch.} & \textbf{F1} & \textbf{Compl.} & $\mathcal{P}$ & \textbf{Arch.} \\
\midrule
\midrule
adult     & 0.768         & 485.2          & 0.5  & 128      & \textbf{0.781} & \textbf{273.2}          & 0.0  & 64  \\
bank      & 0.718         & \textbf{516.0} & 0.0  & 128      & \textbf{0.731} & 610.4          & 0.0  & 128 \\
connect4  & \textbf{0.586} & 7687.4        & 0.9  & $256 \times 3$  & 0.580          & \textbf{2302.6} & 0.0 & 256 \\
letRecog  & \textbf{0.810} & 7334.6        & 0.9  & $256 \times 3$  & 0.655          & \textbf{1779.0} & 0.5 & 256 \\
magic04   & \textbf{0.795} & 921.4         & 0.0  & 256      & \textbf{0.794}          & 894.4           & 0.0 & 256 \\
mush. & \textbf{1.000} & 42.8          & 0.0  & 128      & \textbf{1.000} & \textbf{42.6}   & 0.0 & 64  \\
wine      & \textbf{0.932} & 39.6          & 0.0  & 64       & 0.906          & \textbf{34.4}   & 0.0 & 128 \\
\bottomrule
\end{tabular}%
}
\label{tab:table}
\end{table}

\subsection{Limitations}

{It should be noted that the method does not offer any guarantees on the stability of the rule set: as shown by the variability in performance (Fig.~\ref{fig:repl_and_l0_connect_4_f1_crs}) and literals (Fig.~\ref{fig:rulesvsf1}), it is possible to end up with different models, as it is usually the case for rule-set based models~\cite{BERGAMIN2025128699}; in the cited work, Bergamin et al. also provide methods to improve the robustness and stability of these algorithms through the aggregations of multiple independent runs.  Another important point involves the feature representation: the method is constrained to work with only the binary features provided, other than aggregating binary features through the use of AND and OR operations. %
}

\section{Conclusions and Future Works}\label{sec:conclusion}

In this work, we proposed an adaptation of a computationally complex regularization technique into the MLLP framework, a {network based on propositional logic} that acts as the differentiable version of a multi-level hierarchical rule-based model.
We enhanced the interpretability of rule-based models by reducing their complexity through a model compression technique leveraging a regularizer during the optimization. We showed that our integration can reduce model complexity with a limited effect on performance, which can sometimes be boosted. {This evidence goes against the common belief that trading accuracy for interpretability is needed: in fact, existing literature shows that promoting simpler models is prone to less overfitting and improved robustness \cite{Rudin2018StopEB}. }
In future work, we plan to introduce the inclusion of logical constraints in the network. Specification of data requirements through explicit background knowledge could help the network meet desirable properties such as safety and fairness.

\begin{table}[t]
    \centering
    \vspace{-1.25em}
        \caption{Sample models for the mushroom dataset.}
        \vspace{-1.5em}
\resizebox{0.75\columnwidth}{!}{%
    \begin{tabular}{c p{8.5cm}}
    \multirow{-1}{*}{\rotatebox[origin=c]{90}{Baseline model $\quad \quad \quad \; \;\;\;\;$ }} & 
\begin{empheq}[box=\widefbox]{align*}
    {\text{cap-color} = y \wedge \text{gill-size} = n \wedge \text{stalk-surface-above-ring} = y\; \wedge} \\
    {\text{stalk-surface-below-ring} = y \wedge \text{stalk-color-above-ring} = y \; \wedge} \\
    {\text{stalk-color-below-ring} = y \wedge \text{veil-color} = y}  
\end{empheq}
    \[
\begin{aligned}
    &\vee \, \boxed{\text{gill-size} = n \wedge \text{stalk-shape} = e \wedge \text{stalk-root} = b \wedge \text{ring-type} = p} \\
    &\vee \, \boxed{\text{gill-spacing} = c \wedge \text{gill-size} = n \wedge \text{spore-print-color} = w} \\
    &\vee \, \boxed{\text{gill-spacing} = c \wedge \text{stalk-surface-above-ring} = k} \\
    &\vee \, \boxed{\text{spore-print-color} = r} \, \vee \, \boxed{\text{odor} = p} \, \vee \, \boxed{\text{odor} = f} \\
\end{aligned}
\]
\\
\hline
    \multirow{-1}{*}{\rotatebox[origin=c]{90}{Regularized model  $ \quad \,$  }} & 
    \[
\begin{aligned}
    &\boxed{\text{bruises} = t \wedge \text{gill-size} = n \wedge \text{stalk-shape} = e} \\
    &\vee \, \boxed{\text{gill-spacing} = c \wedge \text{stalk-surface-above-ring} = k} \\
    &\vee \, \boxed{\text{gill-size} = n \wedge \text{population} = c} \\
    &\vee \, \boxed{\text{odor} = c \wedge \text{gill-size} = n} \\
    &\vee \, \boxed{\text{spore-print-color} = r} \, \vee \, \boxed{\text{gill-color} = b} \, \vee \, \boxed{\text{odor} = f}
\end{aligned}
\]

    \end{tabular}%
   }
        \vspace{-2.5em}
    \label{tab:example}
\end{table}

\vspace{-0.2em}
\section*{Acknowledgements}

\scriptsize{
Roberto Confalonieri and Fabio Aiolli acknowledge the partial support of the European Union under the National Recovery and Resilience Plan (NRRP), Mission 4 Component 2 Investment 1.3 -- Call for tender No. 341 of March 15, 2022 of Italian Ministry of University and Research – NextGenerationEU; Code PE0000013, Concession Decree No. 1555 of October 11, 2022 CUP C63C22000770006, project title: "Future AI Research (FAIR) -- Spoke 2 Integrative AI -- Symbolic conditioning of Graph Generative Models (SymboliG)”. Roberto Confalonieri also acknowledges ‘NeuroXAI’ (BIRD231830) funding. 

Finally, the authors would like to thank Gonzalo Jaimovitch-López for his MSc work.
}

\vspace{-0.5em}

\bibliographystyle{IEEEtran}
\bibliography{IJCNN2025/references}

\end{document}